\documentclass[sigconf]{acmart}

\AtBeginDocument{%
  }

\usepackage{enumitem}

\setcopyright{acmlicensed}
\copyrightyear{2026}
\acmYear{2026}
\acmDOI{}
\acmConference[RL-Eval '26]{Workshop on RL for Evaluation}{2026}{San Jose, CA, USA}
\acmISBN{}

\begin{document}

\title{Beyond Static Evaluation: Building Simulation Environments for Scalable Agentic Reinforcement Learning}
\author{Akshay Arora, Ishan Nigam, Ashutosh Aggarwal, Shefali Bansal, Krishna Singh\\Sweta Kumari, Nikhil Mittal, Shariq Farhan, Siddarth Malreddy}
\affiliation{%
  \institution{Uber AI Solutions}
  \city{San Francisco}
  \state{CA}
  \country{USA}
}
\email{smalreddy@uber.com}

\renewcommand{\shortauthors}{Arora et al.}

\begin{abstract}
As Large Language Models (LLMs) evolve into autonomous agents, traditional static evaluation fails to capture multi-step decision-making. We introduce \textit{AgenticAI-Supervisor}, an API and UI-driven RL Gym environment that decouples environment creation from scalable execution. By moving to verifiable execution outcomes, the platform generates high-fidelity traces and applies multi-dimensional reward shaping. Critically, our framework mitigates reward hacking through rigorous internal state validation and testing. This work provides a first look at our platform's core capabilities through a Customer Support Agent case study demonstrating a consistent closed-loop feedback for model optimization. Future work will focus on advanced features such as Computer Use, Tool Use, automated "stumping", and edge-case generation.
\end{abstract}

% CCS Concepts and Keywords go here

\maketitle

\section{Introduction}
\label{sec:intro}
Large Language Models (LLMs) are transitioning from conversational interfaces into autonomous agents capable of reasoning across external tools and complex GUI-based applications~\cite{Shah2026, ReliabilityBench2026, Zhu2025}. Unlike traditional chatbots, these agents operate in dynamic environments where success depends on long-horizon planning and error recovery~\cite{Mohammadi2025}. However, as agentic workflows expand, static single-turn benchmarks fail to capture the multi-step decision-making and environmental feedback these agents require~\cite{ReliabilityBench2026}. Consequently, enterprise-grade models exhibit a severe reliability gap, failing approximately 76\% of complex professional tasks due to compounding execution errors~\cite{Xu2025, Shah2026}. In high-stakes operations such as supply chain auditing or procurement, enterprises cannot rely on systems prone to losing logical consistency or violating implicit constraints over long horizons~\cite{Tau2Bench2026}.

\begin{figure}[t!]
  \centering
  \includegraphics[width=\linewidth]{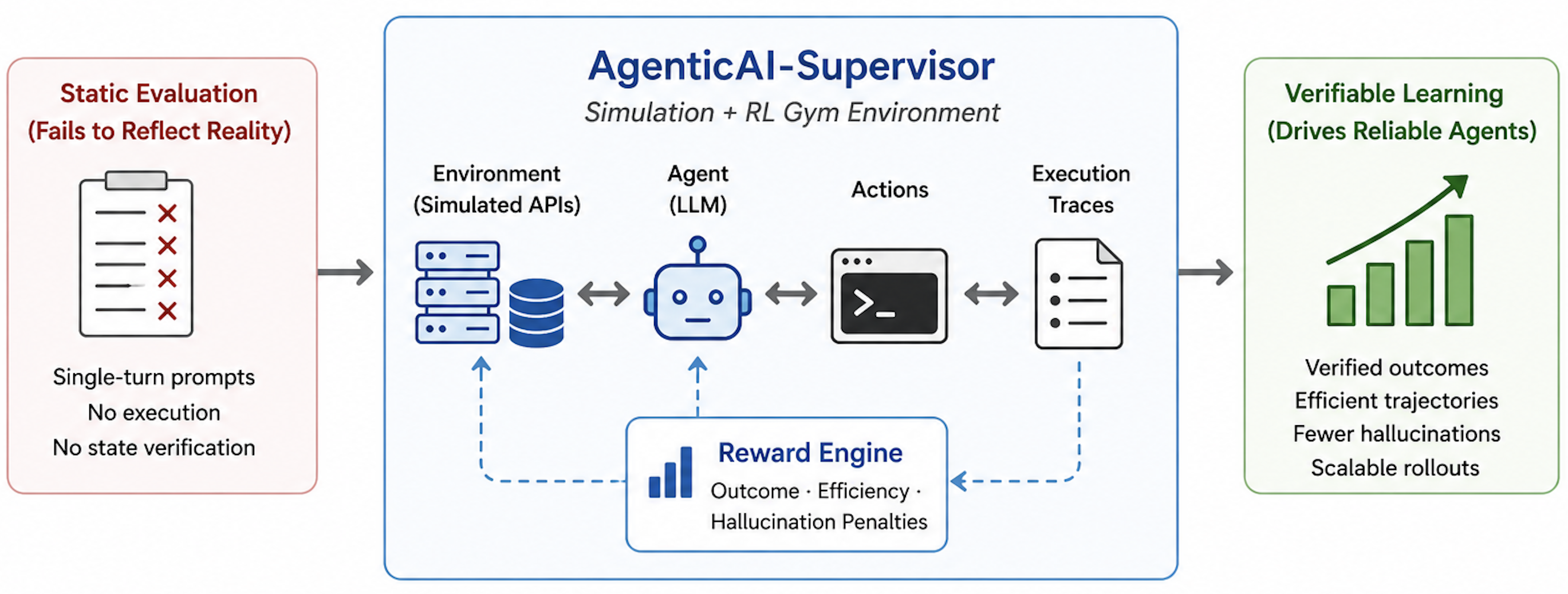}
  \caption{AgenticAI Supervisor enables scalable, verifiable reinforcement learning by combining simulated environments, execution traces, and multi-dimensional rewards.}
  \label{fig:my_single_fig}
  \Description{A text description of the figure for screen readers. This is required by the ACM template.}
\end{figure}

To establish trust in autonomous systems, the evaluation paradigm must shift from grading textual responses to verifying programmatic actions within large-scale reinforcement learning (RL) lifecycles~\cite{Zheng2026}. While execution traces enable deep-dives into reasoning failure modes~\cite{Lee2026}, the manual authorship of multi-step test cases, ``stumping'' prompts, and edge scenarios remains an error-prone bottleneck~\cite{Gao2025}. This creates a scaling deficit, necessitating foundational infrastructure that can automate high-fidelity environment generation and bridge the gap between subjective heuristics and verifiable state validation. 

To address these challenges, we introduce the \textit{AgenticAI-Supervisor}, a foundational API and UI-driven simulation environment engineered for continuous agent evaluation and optimization \cite{Muller2026}. Operating as a high-fidelity ground-truth engine, the platform enables thousands of parallel, isolated trajectories within a secure sandbox where agents interact with simulated tools and web-based UIs \cite{Zhang2026, Muller2026}. By integrating simulation directly into the authoring pipeline, we establish a scalable ``Run-to-Verify'' loop that ensures 100\% data integrity and generates the verifiable rewards necessary for deep RL \cite{Zhang2026}. Our framework specifically mitigates reward hacking through rigorous internal state-mutation testing, ensuring agent trajectories align with business logic rather than exploiting heuristic gaps. Our core contributions are:
\begin{itemize}
    \item A dual-phase framework that decouples synthetic environment scaffolding from high-concurrency rollout execution.
    \item A deterministic reward shaping engine that moves beyond heuristic evaluation to strict state validation, penalizing hallucinations and enforcing trajectory efficiency.
    \item A case study of an autonomous Customer Support agent, demonstrating our efficacy in generating high-fidelity traces for multi-step constraint reasoning and planning.
\end{itemize}

\section{Related Work}

This section reviews recent literature on evaluating and optimizing autonomous agents. We focus on the shift toward interactive simulation environments (Section~\ref{subsec:interactive_envs}), the adoption of Reinforcement Learning (RL) for multi-step workflows (Section~\ref{subsec:rl_workflows}), and the use of automated reward shaping via deterministic and LLM-based verifiers (Section~\ref{subsec:verifiers}).

\subsection{From Static Benchmarks to Interactive Environments}
\label{subsec:interactive_envs}
Historically, LLMs have been evaluated using static, single-turn benchmarks (e.g., MMLU or GSM8K), which measure isolated textual predictions or reasoning capabilities without environmental feedback. However, as LLMs are increasingly deployed as autonomous agents, these static metrics fail to capture the multi-step decision-making, planning, and tool-use required for real-world tasks \cite{Zhu2025, Mohammadi2025}. To bridge this gap, interactive evaluation frameworks and benchmarks such as ALFWorld, WebShop, and WebArena were introduced. These platforms simulate partially observable environments where agents must navigate state changes and adapt to dynamic feedback over time \cite{Wang2024, Xu2025}. The \textit{AgenticAI-Supervisor} extends these paradigms by focusing on high-fidelity, enterprise-specific API emulations rather than generalized web browsing or game-based tasks, providing a verifiable "Run-to-Verify" sandbox.

\subsection{Reinforcement Learning for Agentic Workflows}
\label{subsec:rl_workflows}
While Supervised Fine-Tuning (SFT) is effective for teaching models basic tool-calling syntax and chat templates, it is insufficient for training robust agents capable of error recovery and long-horizon planning \cite{Fireworks2025, Unsloth2026}. Reinforcement Learning (RL) has emerged as the standard for optimizing agentic behavior across entire trajectories. Rather than forcing a model to mimic a static "golden path," modern RL techniques such as Proximal Policy Optimization (PPO) and Group Relative Policy Optimization (GRPO) allow the agent to explore reasoning paths and receive feedback based on verifiable outcomes \cite{Unsloth2026}. A critical challenge in multi-turn RL is environment stabilization and state management; our platform addresses this by executing isolated rollouts via stateless container jobs that interact with deterministic mock databases.

\subsection{Automated Evaluation and Reward Shaping}
\label{subsec:reward_shaping}
A primary bottleneck in scaling RL for agentic systems is the design of reliable reward functions. In closed-loop systems, deterministic verifiers which check for database mutations, verify final costs against budget constraints, or pass unit tests provide the most robust reward signals resistant to reward gaming \cite{Fireworks2025}. However, for open-ended or qualitative interactions, recent frameworks increasingly rely on "LLM-as-a-Judge" mechanisms. These models evaluate trajectories at the turn-level or episode-level to provide scalar rewards or detect major deviations \cite{Tan2026, Chen2025}. Our architecture incorporates both: strict programmatic verifiers for environment state validation and trajectory efficiency, combined with LLM judges to penalize hallucinations and enforce soft constraints.

%%%%%%%%%%%%%%%%%%%%%%%%%%%%%%%%%%%%%%%%
%%%%%%%%%%%%%%%%%%%%%%%%%%%%%%%%%%%%%%%%

\section{Simulation and Execution Framework}
\label{sec:architecture}
Generating high-fidelity reinforcement learning (RL) data requires environments that transition beyond static synthesis to functional real-world representation. To prevent performance divergence between simulation and production, our architecture decouples environment instantiation from large-scale execution. This systemic separation ensures that agent evaluations remain representative of actual operational constraints.

\subsection{High-Fidelity Environment Scaffolding}
\label{subsec:env_scaffolding}
Enterprise workflows involve complex dependencies and non-deter- ministic error paths that are difficult to replicate in static simulations. To capture this functional fidelity, our framework utilizes three core components:
\begin{itemize}
    \item \textbf{Agentic Workflows:} Domain-driven execution paths that incorporate standard operations alongside deliberate failure states, missing data, and ambiguous tool responses to test agent resilience.
    \item \textbf{Base Tool Simulator:} Reusable infrastructure for stateful tools spanning backend APIs and interactive web-based UIs exposed via the Model Context Protocol (MCP)~\cite{mcp_anthropic} specification for tool interface standardization.
    \item \textbf{Dataset Connectors:} A state management layer that binds test cases to specific environmental contexts ensuring consistent initialization and grounding for each rollout.
\end{itemize}

\begin{figure*}[t]
  \centering
  \includegraphics[width=.8\textwidth]{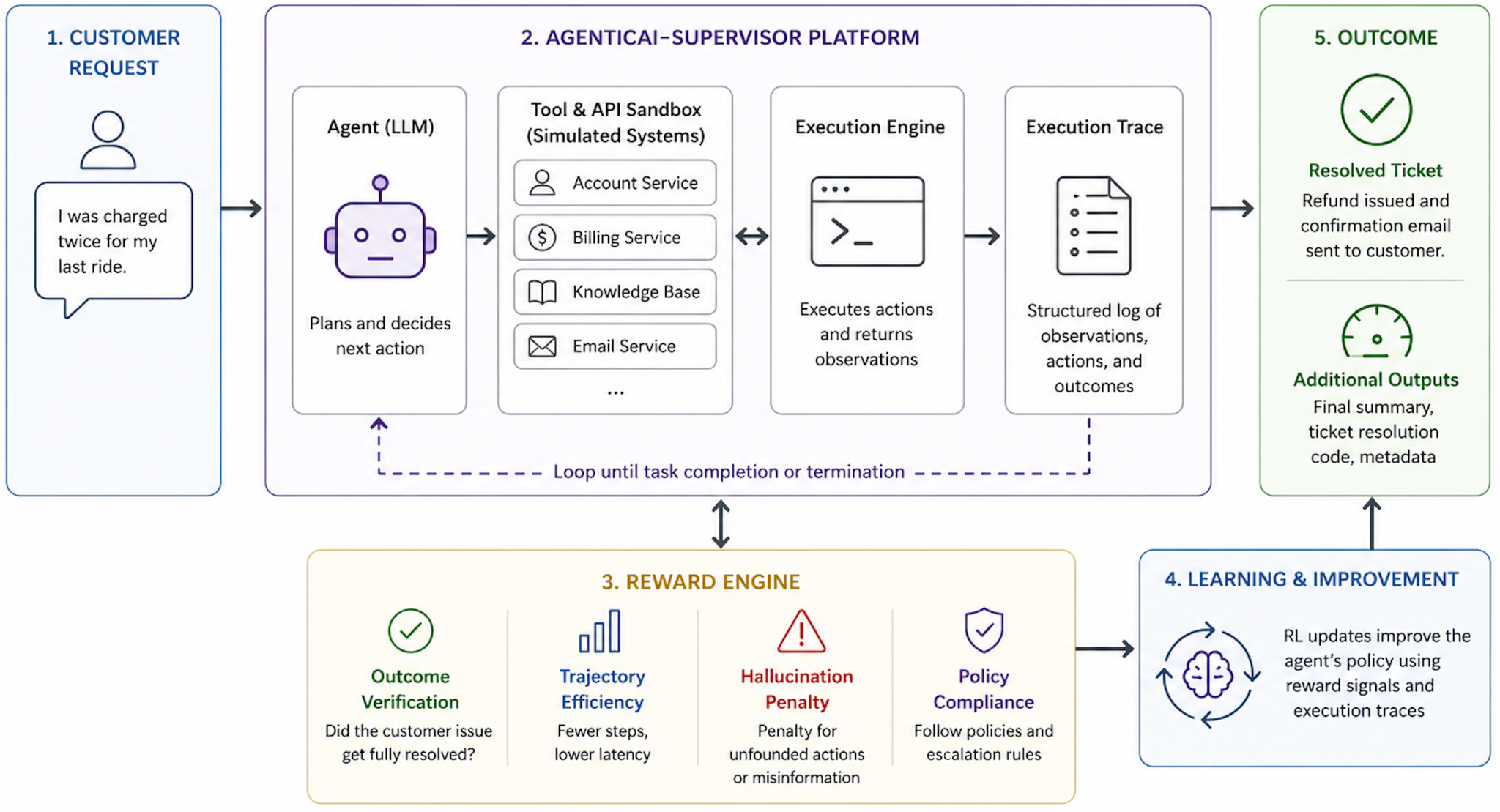}
  \caption{Autonomous customer support built using AgenticAI-Supervisor. A customer request is handled by an LLM agent operating inside a simulated support environment with read-only tools for account, billing, policy, and communication lookup, and actionable tools for resolution workflows. Each interaction is executed inside the sandbox, logged as an execution trace, and scored by a reward engine based on outcome verification, trajectory efficiency, hallucination penalties, and policy compliance. These signals drive iterative reinforcement learning to improve resolution quality and reliability over time.}
  \label{fig:my_wide_fig}
  \Description{A text description of the wide figure for screen readers.}
\end{figure*}

\subsection{Scalable Execution Engine and Rollout Orchestration}
\label{subsec:execution_engine}

The Test Run Engine operationalizes parallel rollouts within isolated, stateless sandboxes to ensure statistical significance. The Rollout Handler provisions containerized instances with a discrete lifecycle, preventing state leakage between iterations. Within the environments, the Agent Runtime orchestrates the interaction loop incorporating LLM prompting, action parsing for tool calls and GUI interactions, and observation retrieval. For observability, discrete events (invocations, executions, reward assignments) are logged as structured \textit{Spans}. Aggregations of these spans form high-fidelity execution \textit{Traces} used for debugging and reward modeling.

%%%%%%%%%%%%%%%%%%%%%%%%%%%%%%%%%%%%%%%%
%%%%%%%%%%%%%%%%%%%%%%%%%%%%%%%%%%%%%%%%

\section{Closed-Loop Reward Formulation for Reinforcement Learning}
\label{sec:reward_formulation}

% Too long... Can't spend 1+ page on this...!
% Text seems quite complex. Consider writing in as simple a language as possible. 
% References to external concepts such as TraceStore are confusing for manuscript readers who don't have any other context.
% Adding equations without defining terms is a sin. For example, $C_{\text{total}} \leq B_{\text{expected}}(1+\tau)$ doesn't explain either the terms or the purpose beyond what is said in plain English.
% Also, the first three equations in Trajectory Efficiency Reward just restate in math what was said in English. The last two did require equations but then did not use them properly.

The reward framework combines terminal verification criteria with a continuous trajectory efficiency signal, providing both sparse task-completion feedback and dense behavioral quality signal for policy optimization.

\subsection{Execution Trace Repository}
\label{subsec:trace_gen}

All rollout data including reward scores, per-criterion subscores, and  intermediate spans is committed upon episode completion. Each span encodes a discrete event (LLM invocation, tool call, state change, reward assignment). Their aggregation forms a complete execution trace for offline debugging and batch sampling.

\subsection{Multi-Dimensional Reward Shaping}
\label{subsec:m_reward_shaping}

The reward framework targets three dimensions, each addressing a distinct failure mode observed in deployed agentic systems. Terminal criteria such as \textbf{Outcome Reward} and \textbf{Constraint Adherence} supply sparse binary signals on whether the agent reached a valid final state. These are complemented by a continuous \textbf{Trajectory Efficiency Reward} component that provides dense process-based feedback spanning tool correctness, call redundancy, API error rate, required-tool coverage, and step economy.\\

\noindent\textbf{Outcome Reward} The terminal environment state is compared against curated \emph{golden answers} via multiset equality over normalized action keys. Additionally, the committed state must satisfy a configurable pre-determined resource budget. % $C_{\text{total}} \leq B_{\text{expected}}(1+\tau)$.
This primary task-completion signal is binary and is blind to the agent's natural language output and trajectory quality.\\

\noindent\textbf{Constraint Adherence} Agents may satisfy outcome criterion through strategies that violate implicit behavioral constraints. Real-world analysis (detailed in Sec. 5) found constraint misrepresentation in ${\sim}40\%$ of positively reinforced episodes and fabrication in ${\sim}3.8\%$ of episodes under outcome only rewards, motivating explicit constraint verification. Constraint Adherence enforces \textbf{(i)} negative checks: records must not assume forbidden field values, \textbf{(ii)} side-effect detection: entity counts at episode end are compared against setup-time baselines catching reward hacking via spurious record creation, and \textbf{(iii)} output fidelity: fabricated factual claims in the agent's response are cross-referenced against tool API responses in the trace.

\noindent\textbf{Trajectory Efficiency Reward} This is the primary process-based signal in our reward formulation and evaluates behavioral quality by directly using the structured tool calls, GUI events, and result records. It consists of the following five sub-components:

\begin{itemize}[leftmargin=1.5em, itemsep=4pt]

\item \emph{Tool Correctness}: %($R_{\text{tool}}$). 
Individual tool invocations are scored for validity independently of overall trajectory shape. The reward linearly combines the fraction of valid (non-error, non-banned) calls with a penalty for invoking explicitly prohibited actions.
%: 
%\begin{center}
%\vspace*{-4mm}
%$R_{\text{tool}} = \alpha \cdot \text{valid\_calls\_ratio} - \beta \cdot \text{banned\_call\_violations} \label{eq:rtool}$
%\end{center}

\item \emph{Redundant Call Penalty}: %($P_r$).
A tool invocation is redundant if the same tool is called with identical parameters after a prior successful invocation already exists in the trace, inflating trajectory length without adding information.
%:
%\begin{center}
%\vspace*{-4mm}
%$P_r = \text{redundant\_calls} / \text{total\_tool\_calls}$
%\end{center}

\item \emph{Validation Error Penalty}: %($P_e$).
The proportion of tool invocations returning an error response, penalizing malformed parameters and invalid API usage.%:
%\begin{center}
%\vspace*{-4mm}
%$P_e = \text{error\_calls}/\text{total\_tool\_calls}$
%\end{center}

\item \emph{Min-Tool Coverage Score}: Ensures that the required tools are invoked according to task specifications, penalizing both under-calling and excessive usage. A smooth non-linear function blends the deficit and excess penalties, which is then sigmoid-aggregated with redundancy and error penalties to produce a continuous unit-normalized trajectory efficiency score.

%($S_m$).
%Verifies that each required tool was invoked at least a minimum number of times per the task specification. Both deficit (under-calling) and excess (over-calling beyond configurable slack $\varepsilon$) are penalized using a smooth non-linear function, avoiding the discontinuous gradient cliffs of hard thresholds.%:
%\begin{align*}
%  d_i &= \frac{\max(0,\, r_i - n_i)}{\max(1,\, r_i)} \\[4pt]
%  e_i &= 1 - \exp\!\left(-\alpha \left(\frac{\max(0,\, n_i - r_i - \varepsilon)}{\max(1,\, r_i)}\right)^{\!2}\right) \\[4pt]
%  S_m &= 1 - \operatorname{mean}_i\!\bigl[\delta\, d_i + (1-\delta)\, e_i\bigr]
%\end{align*}
%where $r_i$ is the required call count for tool $i$, $n_i$ the observed count, $\alpha$ controls penalty curvature, and $\delta$ blends deficit and excess penalties. The sub-components are sigmoid-aggregated into a continuous score in $[0,1]$:
%\vspace*{-2mm}
%\begin{equation*}
%R_{\text{traj}} = \sigma\!\left(w_m S_m - w_r P_r - w_e P_e\right)
%\label{eq:rtraj}
%\end{equation*}

\item \emph{Step-Penalized Efficiency Modifier}: Penalizes unnecessarily long rollouts by scaling the reward by the difference between the executed and ground truth traces. A configurable unit-bound hyperparameter controls the rate of decay providing a differentiable verbosity penalty. This incentivizes agents to reach correct terminal states via minimal reasoning chains.% aligning with efficiency signals utilized in recent multi-turn RL research.

%($M_{\text{eff}}$). To penalize unnecessarily long rollouts without hard cutoffs, a multiplicative modifier scales the terminal reward by an exponential decay over the gap between observed and optimal step counts:
%\begin{equation}
%  M_{\text{eff}} = \alpha^{\,n_{\text{actual}} - n_{\text{optimal}}}
%  \label{eq:meff}
%\end{equation}
%where $\alpha \in (0,1)$ controls per-step decay.
%$M_{\text{eff}}$ provides a smooth, differentiable verbosity penalty analogous to step-penalized reward efficiency signals in recent multi-turn RL work, incentivizing agents to reach correct terminal states via minimal reasoning chains.

\end{itemize}

\subsection{Deterministic and LLM-as-Judge Verifiers}
\label{subsec:verifiers}

Evaluation is implemented through two modalities. \textbf{(1) Verifiable Rewards} perform deterministic, state-based checks such as matching golden answers, validating environment constraints, and cross-referencing identifiers across API and GUI traces without incurring inference costs. Qualitative dimensions, including response coherence and reasoning quality, are assessed via \textbf{(2) LLM-as-a-Judge} evaluators using structured rubrics. This modality provides a dense partial-credit signal based on the fraction of essential criteria met, rather than a binary pass/ fail evaluation. To ensure consistency, we employ ensemble judging to reduce evaluator variance with the final reward balance being calibrated via configuration to suit specific environment requirements.

%Reward Connectors implement two evaluation modalities: \texttt{(1)}  \textbf{Verifiable Rewards} handle structured, state-based assessments such as golden answer matching, constraint checks, identifier and cost cross-referencing deterministically and without inference cost, and \texttt{(2)} \textbf{LLM-as-a-Judge} evaluators handle qualitative dimensions that resist deterministic verification—response coherence, policy-tone adherence, and reasoning quality—via structured rubrics specifying \emph{essential} and \emph{optional} criteria. A dedicated \emph{reasoning quality} rubric ($R_{\text{cot}}$) targets the chain-of-thought specifically: evaluating logical consistency, step-by-step planning justification, and the absence of self-contradictory inference. The reward is the fraction of essential criteria passed ($R_{\text{llm}} = k_{\text{pass}} / K$), yielding partial-credit signal denser than a binary rubric pass. Ensemble judging across multiple invocations reduces evaluator variance. The contribution of each modality is governed by the reward configuration, enabling per-environment calibration without changes to core grading infrastructure.

\section{Case Study: Autonomous Customer Support}
\label{sec:case_study}

To validate the simulation environment's efficacy, an Agentic AI Reinforcement Learning Environment was designed around a Customer Support domain, tasking the agent with autonomously resolving customer issues, processing transactions, and managing account security.

\subsection{Emulated Domain Tools and API Integration}
The emulated environment features a bifurcated suite of API and GUI tools, categorized into "Actionable" and "Non-Actionable" (Read-Only) tools. This structure necessitates that the agent gathers context before executing state-mutating actions.

The non-actionable toolkit provides observational capabilities. The agent utilizes \texttt{get\_customer\_info} and \texttt{get\_order\_details} to retrieve comprehensive customer profiles and order histories. To ensure consistent support and policy compliance, the agent can verify past communications via \texttt{check\_interaction\_history} and cross-reference company guidelines using \texttt{search\_kb\_and\_policies}.

Once sufficient context is gathered, the agent invokes actionable tools to mutate the environment state:
\begin{itemize}
\item \textbf{Financial \& Fulfilment Tools:} The \textbf{Refund Tool} processes monetary returns, while the \textbf{Replacement Tool} %creates free orders for defective items;
resolves defective item reports by creating a new fulfillment order; both automatically update order statuses and log actions to the associated ticket.
\item \textbf{Security Tool:} Locks accounts for identified fraud (e.g., \texttt{account\_takeover} or \texttt{return\_fraud}), cancels all pending orders, and creates an audit trail.
\item \textbf{Workflow Management:} Tools such as \textbf{create\_ticket} and \textbf{update\_order\_status} allow the agent to manage the lifecycle of customer grievances and order logistics.
\end{itemize}

\subsection{Evaluation Suite and Scenario Complexity}
Evaluation was conducted against a curated suite of tasks designed to test the agent's ability to sequence these tools effectively. Complex scenarios required the agent to cross-reference knowledge base policies before authorizing refunds, or to detect suspicious activity in the interaction history to trigger appropriate security locks. By generating continuous, verifiable reward signals across these simulations, the platform demonstrates its capacity to reinforce safe, policy-compliant customer support workflows.

%\section{Empirical Case Study: Autonomous Travel Planner}
%\label{sec:case_study}
%
%To validate the simulation environment's efficacy, an Agentic AI Reinforcement Learning Environment was designed and deployed for Anthropic. This "Trip Planner" Environment serves as a rigorous domain for evaluating advanced AI agents on multi-step decision-making, long-horizon planning, and constraint reasoning.
%
%The primary objective for the agent is to process a natural language user request and synthesize a comprehensive, actionable travel itinerary. The emulated application environment was provisioned with a suite of domain-specific APIs, including tools to: \texttt{Fetch Nearest Airport}, \texttt{Search Flight by Airport Code}, \texttt{Search Hotels}, \texttt{Search Points of Interest}, and \texttt{Get Weather Details}.
%
%Evaluation was conducted against a curated suite of 100 evaluation tasks, deliberately structured to encompass varying levels of complexity—from simple single-dataset retrievals to intricate tasks requiring constraint handling and multi-dataset planning. For each test case, the final performance score was calculated as the arithmetic mean of the automated grading mechanism (yielding a continuous value between 0.0 and 1.0) aggregated over 10 independent execution runs. For instance, a test case generating scores of 0.1, 0.15, etc., across its 10 attempts results in a stable, averaged reward signal, demonstrating the platform's capacity to provide consistent, closed-loop feedback for model evaluation.

\section{Discussion and Future Work}
\label{sec:discussion}

We outline a roadmap to transition from foundational infrastructure to a comprehensive ecosystem for autonomous agent optimization.

\subsection{No-Code Simulation Interfaces}
\label{subsec:future_nocode}

Future work will externalize ``Gym Factory'' capabilities via a unified portal for self-serve environment customization and out-of-the-box deployment. This no-code dashboard will allow domain experts to utilize drag-and-drop scenario builders to configure mocked tools, bind datasets, and define reward strategies. This will significantly reduce the lead time for deploying specialized RL gyms.
%A primary focus of future work is externalizing the "Gym Factory" capabilities to clients by integrating creation tools directly into a unified portal portal for self-serve, on-demand services for customizing gyms, with out-of-the-box (OOTB) gyms also available for immediate deployment. This transition will feature a no-code dashboard designed for non-technical users and domain experts. By utilizing a drag-and-drop scenario builder, users will be able to effortlessly configure mocked tools, bind custom datasets, and establish reward strategies without writing boilerplate code, thereby significantly reducing the lead time required to deploy specialized RL gyms.

\subsection{Human-in-the-Loop Reward Overrides}
\label{subsec:future_hitl}
While deterministic verifiers and LLM-as-a-judge mechanisms provide scalable reward signals for well-defined tasks, human oversight remains essential for nuanced, high stakes, or consequential decisions. To address this, we plan to integrate Human-in-the-Loop (HITL) workflows, allowing experts to override automated rewards with qualitative feedback directly in the training pipeline. Additionally, an expert marketplace will validate synthesized environments for realism and solvability prior to large-scale deployment.
%Although deterministic verifiers and LLM-as-a-Judge mechanisms provide scalable reward signals, highly nuanced or subjective tasks benefit from human judgment. To address this, the platform will integrate comprehensive Human-in-the-Loop (HITL) workflows. Within the proposed dashboard, domain experts will have the capability to override automated rewards, providing direct, qualitative feedback into the training pipeline. Furthermore, this expert marketplace will be leveraged to add a secondary layer of validation, ensuring that synthesized environments are rigorously reviewed for realism, correctness, and solvability before they are operationalized at scale.

\subsection{Automated "Stumping"}
\label{subsec:future_stumping}

To refine frontier models, agents must navigate increasingly difficult scenarios. Future iterations will automate ``stumping'' the systematic generation of hard task variants by injecting failure modes, ambiguous states, and missing data. This automation accelerates the creation of robust training datasets, preparing agents for the stochasticity of live production.
%To continuously challenge and refine frontier models, it is necessary to subject agents to increasingly difficult scenarios. Future iterations of the platform will feature specialized "stumping" automation designed to assist test creators in generating harder variants of base tasks. This system will systematically expand standard scenarios into complex edge cases, injecting deliberate failure modes, ambiguous states, and missing data. By automating the generation of these high-value "stumping" prompts, the platform will accelerate the creation of robust training datasets, ensuring agents are prepared for the unpredictable nature of live production environments.

\subsection{Uncertainty-Aware Reward Signals}
\label{subsec:future_uncertainty}

A promising direction involves incorporating \emph{semantic entropy} into the reward pipeline to quantify response unreliability without supervision. By measuring divergence across multiple rollouts, high-entropy states can serve as penalty signals or curriculum markers to prioritize complex tasks. This leverages our trace-based infrastructure to identify inconsistent action distributions. Overall, the goal is a platform that autonomously scales environments with minimal effort, establishing a foundation for reliable enterprise agents.
%A promising direction is incorporating \emph{semantic entropy} into the reward pipeline. By sampling multiple rollouts for the same task and measuring divergence in the induced environment-state distributions, semantic entropy quantifies response unreliability without requiring ground-truth labels per rollout. High-entropy states—where the agent's action distribution is broad and inconsistent across rollouts—can serve as a continuous penalty signal or as a curriculum signal to prioritize harder tasks in RL batch construction. This capability is a natural extension of the trace-based infrastructure described in Section~\ref{subsec:trace_gen} and is reserved for future work.

%Ultimately, the goal is to cultivate a platform that can autonomously generate, validate, and scale environments and test cases with minimal engineering effort, establishing a foundational layer for building the next generation of reliable enterprise agents.

\section{Conclusion}
\label{sec:conclusion}
As models transition into enterprise agents, static evaluation is no longer sufficient; what matters is whether an agent can act successfully inside a realistic environment. By standardizing trace-based evaluation and verifiable rewards through the \textit{AgenticAI-Supervisor}, we provide the essential training ground where agents can act, fail, and improve in a secure sandbox before ever touching production data.  Our framework specifically mitigates reward hacking by ensuring that terminal rewards are tied to internal state validation rather than surface-level textual heuristics. While this paper provides a first look at our foundational architecture, future work will focus on sharing more advanced features - such as automated "stumping" - and integrating human-in-the-loop (HITL) workflows with self-serve, no-code UI capabilities to democratize the rapid creation of custom RL gyms.

\bibliographystyle{ACM-Reference-Format}
\bibliography{refs}

\end{document}